\documentclass[10pt,twocolumn,letterpaper]{article}

\usepackage[pagenumbers]{wacv} 

\usepackage{graphicx}
\usepackage{booktabs}

\usepackage{times}
\usepackage{epsfig}
\usepackage[table]{xcolor}
\usepackage[accsupp]{axessibility}

\usepackage{amsmath, amsfonts, amssymb}       
\usepackage{nicefrac}       
\usepackage{microtype}      
\usepackage{multirow}
\usepackage{bm}

\usepackage[pagebackref,breaklinks,colorlinks]{hyperref}

\usepackage[capitalize]{cleveref}
\crefname{section}{Sec.}{Secs.}
\Crefname{section}{Section}{Sections}
\Crefname{table}{Table}{Tables}
\crefname{table}{Tab.}{Tabs.}

\def\wacvPaperID{294} 
\def\confName{WACV}
\def\confYear{2024}

\begin{document}

\title{Exploiting CLIP for Zero-shot HOI Detection Requires \\ Knowledge Distillation at Multiple Levels}

\author{Bo Wan \qquad \qquad Tinne Tuytelaars\\
ESAT, KU Leuven \\
{\tt\small \{bwan,tinne.tuytelaars\}@esat.kuleuven.be}
}

\maketitle

\vspace{-2mm}
\begin{abstract}
\vspace{-2mm}
In this paper, we investigate the task of zero-shot human-object interaction (HOI) detection, a novel paradigm for identifying HOIs without the need for task-specific annotations. 
To address this challenging task, we employ CLIP, a large-scale pre-trained vision-language model (VLM), for knowledge distillation on multiple levels.
Specifically, we design a multi-branch neural network that leverages CLIP for learning HOI representations at various levels, including global images, local union regions encompassing human-object pairs, and individual instances of humans or objects. To train our model, CLIP is utilized to generate HOI scores for both global images and local union regions that serve as supervision signals. The extensive experiments demonstrate the effectiveness of our novel multi-level CLIP knowledge integration strategy. Notably, the model achieves strong performance, which is even comparable with some fully-supervised and weakly-supervised methods on the public HICO-DET benchmark. Code is available at \href{https://github.com/bobwan1995/Zeroshot-HOI-with-CLIP}{https://github.com/bobwan1995/Zeroshot-HOI-with-CLIP}.
\end{abstract}
\vspace{-3mm}
\section{Introduction}
\vspace{-1mm}
HOI detection aims to identify triplets of $\langle$\textit{human, object, interaction}$\rangle$ within the context of a given image, which requires localization of human and object regions and recognition of their interactive behavior, e.g., play-basketball. It enables the intelligent system to understand and interpret human behavior in real-world scenarios, thus playing an instrumental role in anomalous behavior detection~\cite{liu2018ano_pred,pang2020self}, motion tracking~\cite{MRABTI2019145, Nishimura2021sdof} and visual scene understanding~\cite{Peyre_2019_ICCV,Li_2021_CVPR}.

\begin{figure}
    \centering
    \includegraphics[width=0.9\linewidth]{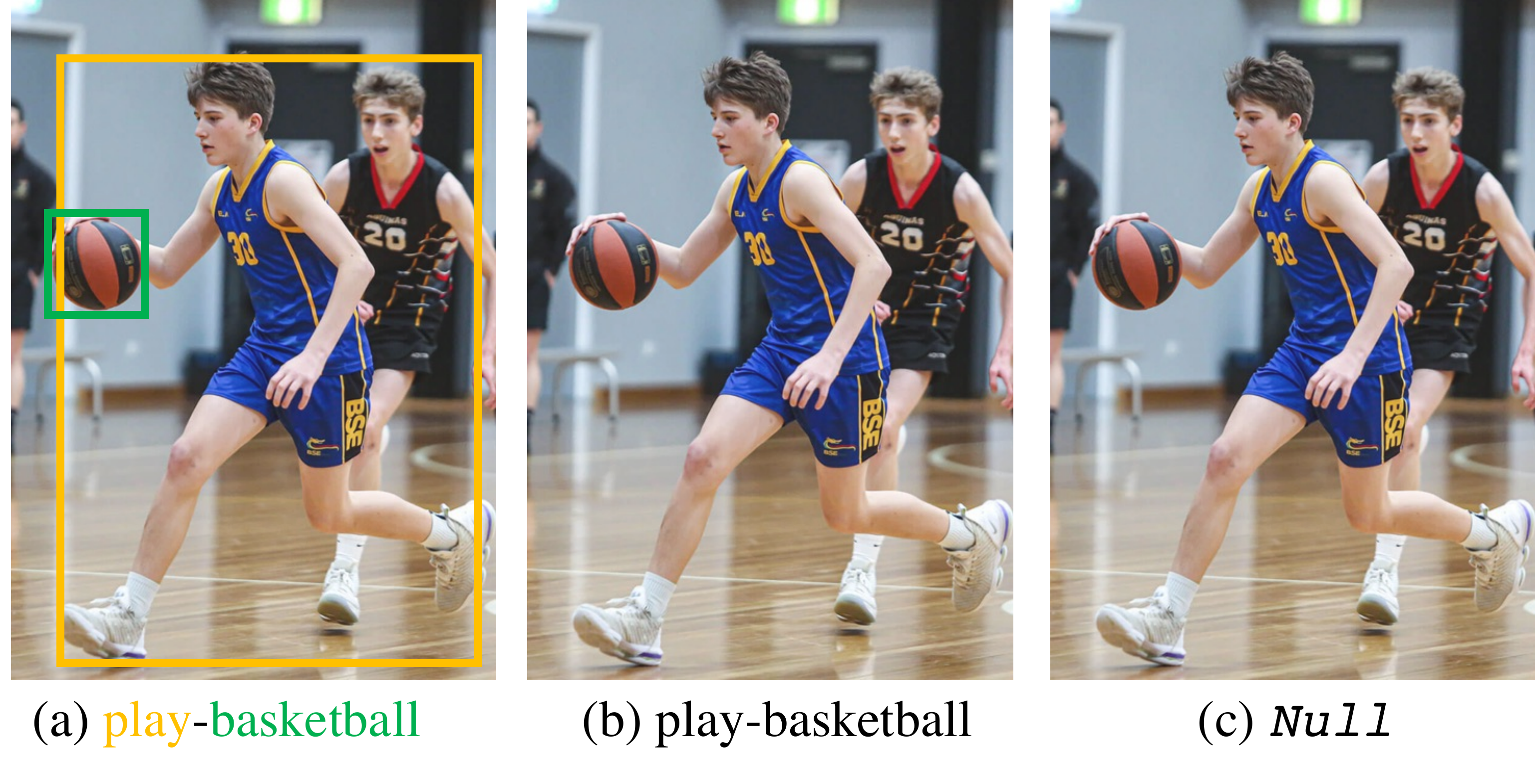}
    \caption{\small{\textbf{Comparison on training annotations for different setups:} (a) Fully-supervised; (b) Weakly-supervised; (c) Zero-shot.
    }}
    \vspace{-4mm}
    \label{fig:ads}
\end{figure}

HOI detection has typically been investigated in a fully-supervised learning paradigm~\cite{chao2018learning,Gao-ECCV-DRG,gupta2015visual,Wan_2019_ICCV,zhang2021spatially}, where aligned HOI annotations (i.e. human-object locations and interaction types) are provided during the training stage, as illustrated in Fig.~\ref{fig:ads}(a). Despite the high performance due to such comprehensive annotations, it suffers from the labor-intensive process of labeling HOI instances. 
This limitation has led recent studies to transition towards a weakly-supervised setup~\cite{zhang2017ppr,baldassarre2020explanation,kumaraswamy2021detecting,kilickaya2021human,wan2023weakly},
where only image-level HOI categories (without corresponding bounding boxes) are provided for model learning, as demonstrated in Fig.~\ref{fig:ads}(b).
While this setup significantly reduces the labeling costs and enhances robustness to label errors, it nonetheless necessitates image-level annotations on HOI datasets. These are still costly to obtain, often noisy and incomplete. Moreover, annotators may inadvertently introduce bias.
Drawing inspiration from the great success of zero-shot learning ~\cite{alec21clip,jia2021scaling,zhai2022lit,zhong2022regionclip,li2021grounded,zhang2022glipv2}, we introduce the new challenging problem of zero-shot HOI detection, where \textit{NO} HOI annotations are required for model learning, as shown in Fig.\ref{fig:ads}(c). Importantly, note how our setup diverges from previous zero-shot HOI detection frameworks~\cite{shen2018scaling,gupta2019nofrills,bansal2019detecting,liu2020consnet,ning2023hoiclip}, 
which predominantly focus on knowledge transfer from observed HOI concepts to unseen categories (c.f. Sec.~\ref{sec:rw} for more details).
In contrast, this work takes it a step further by tackling the extreme scenario where none of the HOI categories have been annotated during training, although we assume the category space is known. 

Large-scale pre-trained VLM, such as CLIP~\cite{alec21clip}, have shown substantial promise in various domains of zero-shot learning, including image classification~\cite{jia2021scaling,zhai2022lit,shen2022k,li2022supervision,yang2022unified,mu2021slip,chen2022altclip}, object detection~\cite{li2021grounded,zhong2022regionclip}, and instance segmentation~\cite{zhang2022glipv2}. 
However, extending these models to zero-shot HOI detection poses a unique challenge, primarily due to the high-level relational understanding required in this context. 
To the best of our knowledge, this paper makes the first effort to propose and tackle this challenging task.
A naive solution might be directly employing CLIP on union regions (the joint areas of human-object proposals detected by an external object detector) to produce corresponding HOI scores. However, this approach proves to be suboptimal in terms of both inference speed and performance, as illustrated in  Sec.~\ref{ablation}. Alternatively, we leverage the power of CLIP in two ways:
i) In terms of model design, we build a multi-branch network that extracts CLIP-oriented features on multiple levels,
thus incorporating its robust generalization capabilities into HOI representations. ii) For model learning, we use the CLIP scores generated from global images and union regions as supervisory signals to train our multi-branch network.

\begin{figure}
    \centering
    \includegraphics[width=0.9\linewidth]{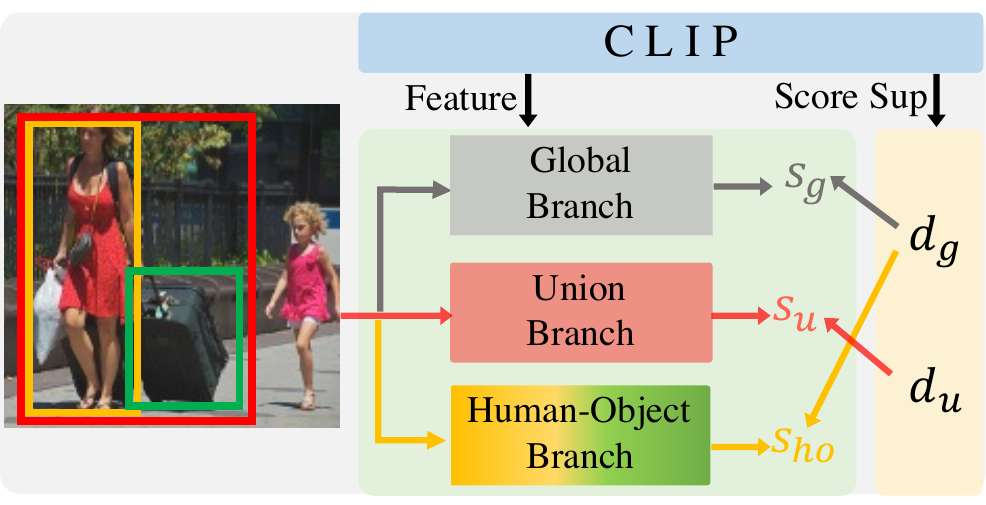}
    \caption{\textbf{Overview of our multi-level knowledge distillation.
    }}
    \vspace{-3mm}
    \label{fig:demo}
\end{figure}

 Prior works~\cite{Wan_2019_ICCV,wu2022end,liu2020consnet} have revealed that HOI relations manifest across various levels, including global images, relational union regions, and individual human-object instances. Inspired by this insight, our work employs a multi-branch neural network to leverage the capabilities of CLIP for multi-level HOI representation learning. As sketched in Fig.~\ref{fig:demo}, the network architecture comprises a global branch, a union branch, and a human-object branch. The process begins with computing an HOI embedding by using the CLIP text encoder to encode HOI label prompts. This embedding is shared across all branches. Concurrently, we detect human-object proposals with an off-the-shelf object detector and generate an image feature map with the CLIP visual encoder. Each branch within our network adopts a similar design, where the HOI features of global image, union regions, and human-object pairs are extracted from the image feature map,
 and then serve to compute scores corresponding to the HOI embedding. 
 Finally, we bring these branches together using a late fusion strategy: rather than focusing on merging semantic features, our approach concentrates on fusing the multi-level HOI scores, offering a comprehensive view of HOIs on different scales.


To generate supervision for model training, we leverage CLIP to produce meaningful HOI scores for both global images and local union regions. For global images, the entire image is re-scaled and fed into the CLIP model to derive a global HOI score that captures the entire context of the image. Given that CLIP is pre-trained to understand a wide range of image-text pairs, it can provide a holistic understanding of visual relationships. Similarly, for union regions, we extract regions of interest that include both the human and the object. These regions, capturing more focused and localized interactions, are separately fed into CLIP to generate local and region-specific HOI scores. 
The local HOI scores can be noisy due to the coexistence of multiple HOIs in the same region or the presence of HOI irrelevant distractions,
but they complement the global supervision signal. We conduct ablative studies on the supervision strategy in Tab.~\ref{hico-aba3}, which demonstrate the application of global supervision on the global \& human-object branch, and local supervision on the union branch, yields the most effective results. 

By incorporating CLIP knowledge on multiple levels for both model design and learning, our approach substantially enhances the zero-shot detection capability in a variety of HOI scenarios.
In summary, our main contributions are three-fold:
\begin{itemize}
\vspace{-1mm}
\item We pioneer the challenging task of zero-shot HOI detection, a new learning setup where no HOI annotations are used during training. 
This is a significant leap forward in the field of HOI detection.
\vspace{-2mm}
\item We propose a multi-level knowledge distillation strategy from CLIP for this task, where we seamlessly incorporate CLIP into the model design for detecting HOIs on different scales, and capture both global and local contexts to provide rich supervision for model training.
\vspace{-4mm}
\item Extensive experiments are conducted to verify the effectiveness of our CLIP integration strategies. Impressively, our method achieves a strong performance even on par with some fully-supervised and weakly-supervised methods on HICO-DET benchmarks.
\end{itemize}

\vspace{-2mm}
\section{Related Works} \label{sec:rw}
\vspace{-1mm}
\paragraph{HOI detection} \textbf{\textit{Fully-supervised HOI detection}} has been the most common setup due to its superior performance. Research in this area generally falls into two categories: two-stage and one-stage frameworks. 
Two-stage methods~\cite{gao2018ican,Wan_2019_ICCV,li2020detailed,Gupta_2019_ICCV,Gao-ECCV-DRG,zhang2021spatially,Ulutan_2020_CVPR, Zhou_2019_ICCV,li2018transferable,zhou2020cascaded} adopt a \textit{hypothesize-and-classify} strategy, which first generates a set of human-object proposals with the off-the-shelf object detector, and then enumerates all possible HOI pairs to classify their interactions.
One-stage methods predict human \& object locations and their interaction types simultaneously in an end-to-end manner, which are currently dominated by transformer-based architectures ~\cite{carion2020endtoend,kim2022mstr,dong2022category,zhang2021mining,zhang2021efficient}.

To decrease the reliance on HOI annotations,
\textbf{\textit{weakly-supervised HOI detection}} is proposed to learn HOIs with only image-level annotations. 
Due to the lack of location annotations, current works in this domain adopts the two-stage framework, and they focus on recognizing HOIs by developing advanced network structures to encode context~\cite{baldassarre2020explanation,kumaraswamy2021detecting} and integrating external knowledge for representation learning~\cite{unal2023weaklysupervised,wan2023weakly}. In this work, we propose a novel zero-shot setup without the need for any manual annotations in HOI detection.

\vspace{-4mm}
\paragraph{Zero-shot HOI detection}
As most HOI classes are distributed in a long-tail manner~\cite{shen2018scaling, gupta2019nofrills,wan2023weakly} due to the inherent compositionality of HOIs~\cite{ning2023hoiclip}, previous works on zero-shot HOI detection~\cite{shen2018scaling,Peyre19Detecting,gupta2019nofrills,liu2020consnet,bansal2020detecting,hou2021fcl,hou2021vcl,hou2021atl,wu2022end,liao2022gen,ning2023hoiclip} aim to distill knowledge from observed HOI concept to unseen classes.
They can be categorized into three scenarios: \textit{unseen object}, \textit{unseen action}, and \textit{unseen combination}. There are mainly two streams of research for solving this problem.
One stream~\cite{shen2018scaling,gupta2019nofrills,bansal2020detecting,hou2021fcl,hou2021vcl} focuses on factorizing the human and object features
by performing disentangled reasoning on verbs and objects, which allows the composition of novel HOI triplets for training and inference. Another stream~\cite{liu2020consnet,wu2022end,liao2022gen,ning2023hoiclip} transfers knowledge from knowledge graphs or pre-trained VLM to recognize
unseen HOI concepts. Despite the substantial success of these approaches in knowledge transfer, they still rely heavily on the base knowledge provided by the seen HOI categories. In contrast, our setup does not require any HOI annotations for learning.

\begin{figure*}
    \centering
    \includegraphics[width=0.95\linewidth]{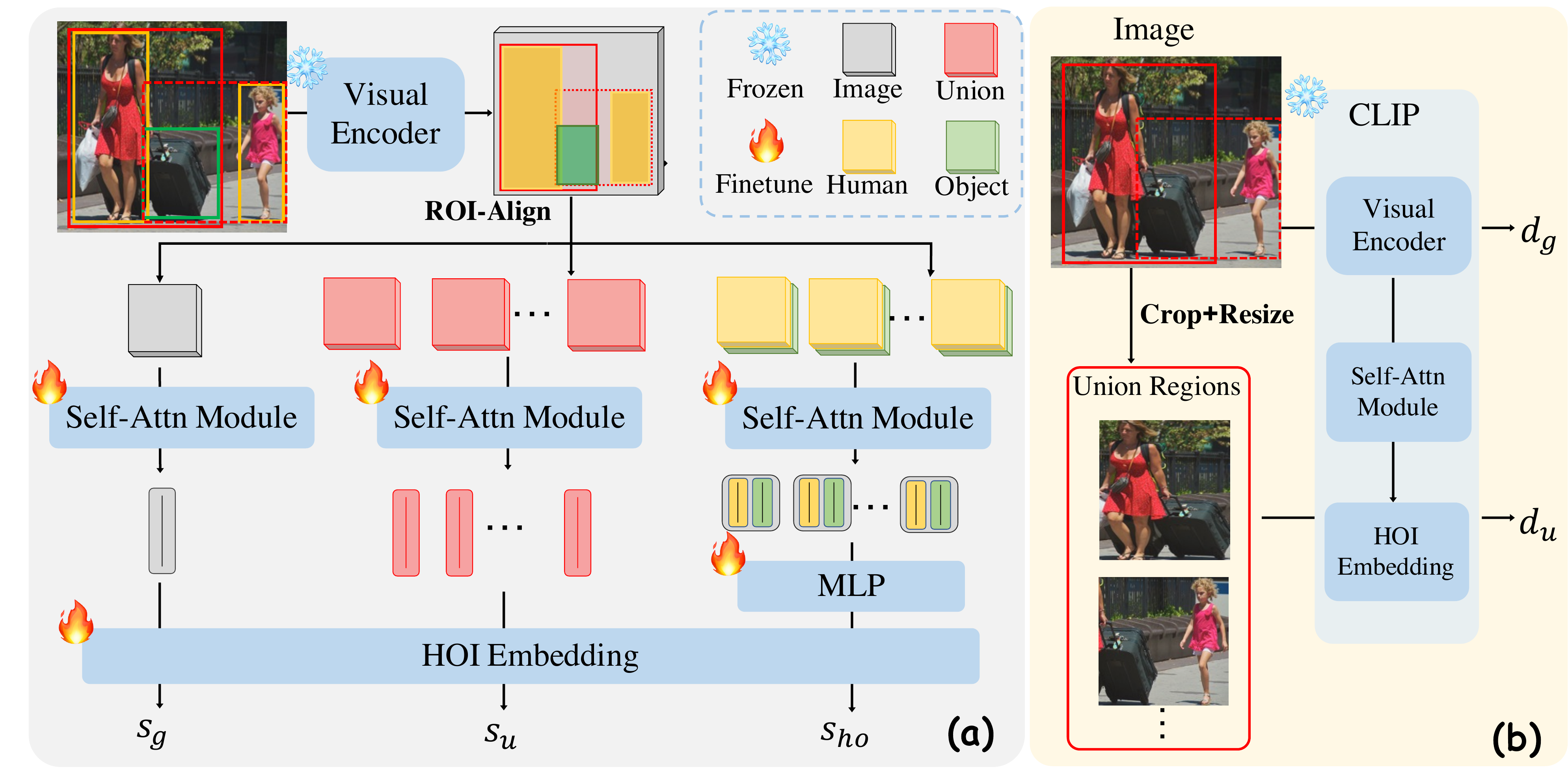}
    \caption{\textbf{A detailed introduction of our method:} We design a multi-branch neural network that incorporates CLIP components to extract multi-level information for HOI detection, which is supervised by the CLIP scores derived from global images and local union regions.
    }
    \vspace{-3mm}
    \label{fig:overview}
\end{figure*}

\vspace{-1mm}
\section{Method}
\vspace{-1mm}
\subsection{Problem Setup}
\vspace{-1mm}
Formally, zero-shot HOI detection aims to learn an HOI detector that takes an image $I$ as input and generates a collection of tuples $ \mathcal{O} = \{ (\mathbf{b}_h, \mathbf{b}_o, r_{h,o}, s_{h,o}^r) \} $. 
Each tuple corresponds to a HOI instance, where $\mathbf{b}_h, \mathbf{b}_o \in \mathbb{R}^4 $ indicate human and object bounding boxes, $r_{h,o} \in \{1,...,N\}$ represents the interaction type 
between $\mathbf{b}_h$ and $\mathbf{b}_o$, and $s_{h,o}^r \in \mathbb{R}$ is the confidence score of the detected interaction.

\vspace{-2mm}
\subsection{Method Overview}
\vspace{-1mm}
To address the challenging zero-shot HOI detection task, we leverage CLIP for multi-level knowledge integration. To this end, we exploit the visual and textual encoders of CLIP to construct a multi-branch network for HOI representation learning, and use CLIP to generate global and local supervision for model training.

\vspace{-3mm}
\paragraph{Model Design} Due to the lack of HOI location annotations, we adopt a typical {\em two-stage} formulation~\cite{zhang2017ppr,baldassarre2020explanation,kumaraswamy2021detecting} for HOI detection: 
in the first stage, we generate a group of human proposals $\{(\mathbf{b}_h, s_h)\}$ and object proposals $\{(\mathbf{b}_o, c_o, s_o)\}$ with an off-the-shelf object detector~\cite{ren2015faster}, where $s_h, s_o \in \mathbb{R}$ are detection scores and $c_o \in \{1,...,C\}$ is the object class. In the second stage, we pair up all human and object proposals and predict the interaction class for each combination.

In order to infuse the generalization capability of CLIP into the HOI representation, we design a multi-branch deep network by incorporating CLIP's visual and textual encoders,
as sketched in Fig.~\ref{fig:overview}. Specifically, the global branch performs image-level HOI recognition, utilizing the HOI embedding produced by CLIP textual encoder as a classifier.
In parallel, for each detected human-object pair $(\mathbf{b}_h, \mathbf{b}_o)$, a union branch extracts the contextual cues in their shared region of interest, providing a comprehensive view of the surrounding environment and potential interactions. On top of that, a human-object branch focuses on fine-grained HOI features and encodes the specific relational attributes of the interactive pairs, which are used to predict their interaction types. All the branches are integrated with a late fusion strategy, where the HOI scores from different levels are combined to obtain the final predictions.

\vspace{-3mm}
\paragraph{Model Learning} To train our model, we first employ CLIP on global image and local union regions to compute the corresponding HOI scores as supervision. Then we apply global image supervision on the global branch and human-object branch, and local union supervision on the union branch. Our training procedure can be viewed as a multi-level knowledge distillation approach from the pre-trained CLIP model. The primary objective of this strategy is to ensure that the HOI scores derived from distinct branches align with the CLIP scores.


\subsection{Model Design}\label{sec:method}

\vspace{-1mm}
\subsubsection{CLIP Backbone}\label{sec:backbone}
\vspace{-1mm}
CLIP builds a powerful vision-language model by pretraining 
on large-scale image-text pairs. 
It consists of a visual encoder $\mathcal{F}_V$ (e.g., a ResNet~\cite{he2016deep} or Vision Transformer~\cite{dosovitskiy2021an}), 
a self-attention module $\mathcal{F}_{ATT}$ and a 
textual encoder $\mathcal{F}_T$ (e.g., a Transformer~\cite{vaswani2017attention}), to map the visual and textual inputs to a shared latent space.

Specifically, for an input image $I$, the 
visual encoder $\mathcal{F}_V$ produces a feature map $\bm{\Gamma} \in \mathbb{R}^{H\cdot W\cdot D}$, 
where $H,W,D$ denote the height, width, and depth of $\bm{\Gamma}$, respectively. 
Then a self-attention module $\mathcal{F}_{ATT}$ is adopted to encode $\bm{\Gamma}$ to a feature vector $v\in \mathbb{R}^D$: it takes a linear projection of the average pooling on the spatial dimensions of $\bm{\Gamma}$ as query $Q \in \mathbb{R}^D$, and a linear projection of the reshaped feature columns as key and value $K,V \in \mathbb{R}^{(HW)\cdot D}$ :
\begin{align}
\vspace{-2mm}
Q = \mathcal{F}_Q(AvgPool&(\bm{\Gamma})); \quad K,V = \mathcal{F}_K(\bm{\Gamma}), \mathcal{F}_V(\bm{\Gamma}) \nonumber\\
v &= \mathcal{F}_{MHA}(Q, K, V)
\vspace{-2mm}
\end{align}
where $\mathcal{F}_{Q,K,V}$ are linear projection layers and $\mathcal{F}_{MHA}$ is a standard multi-head attention module~\cite{vaswani2017attention}, all incorporated in $\mathcal{F}_{ATT}$. 

To leverage CLIP for HOI detection, we utilize CLIP textual encoder $\mathcal{F}_T$ to generate HOI embedding $\mathcal{W}_T\in \mathbb{R}^{N\cdot D}$.
In a prompting strategy akin to CLIP, we adopt a common template
`a person is \{\textit{verb}\}-ing \{\textit{object}\}' to 
convert HOI labels into text prompts. For instance, `play basketball' would be converted to `a person is playing basketball'. These sentences are then processed by $\mathcal{F}_T$ to create the HOI embedding $\mathcal{W}_T$, 
which is used to classify different levels of visual features into corresponding HOI scores.

\vspace{-3mm}
\paragraph{Modified Visual Encoder} 
By default, a CLIP model with a ResNet visual encoder downscales the feature map by a factor of 32.  
Consequently, the spatial dimension of the feature map is insufficient for detailed region-specific feature extraction. 
To adapt this visual encoder for HOI detection with a larger feature map, we tweak the original ResNet structure. 
This is done by discarding the last average pooling module and adding an upsampling layer, thereby reducing the feature map size to only 8 times smaller than the image resolution. To maintain compatibility with the CLIP's self-attention module for feature aggregation, we implement an ROI-Align module\cite{he2017mask} to resize the cropped feature map to a $7\times 7$ grid for global images, union regions, and human-object proposals. 

\vspace{-3mm}
\subsubsection{Global Branch}\label{sec:image_branch}
\vspace{-2mm}
The global branch adapts the visual encoder and HOI embedding to perform an image-wise HOI recognition task, thereby capitalizing on the knowledge embedded within the CLIP model.
Firstly, we adopt an ROI-Align on the image feature map followed by the self-attention module to compute a global feature vector $v_g \in \mathbb{R}^D$.
Then the global HOI scores $s_g\in \mathbb{R}^N$ are predicted by conducting the inner product between the global vector $v_g$ and the HOI embedding $\mathcal{W}_T$: $s_g = \textrm{Softmax}(\mathcal{W}_T \times v_g)$,
where $\times$ is matrix multiplication. 

\vspace{-3mm}
\subsubsection{Union Branch}\label{sec:union_branch}
\vspace{-2mm}
The union region encapsulates the contextual relationship between humans and objects, which is crucial in comprehending their context.
To exploit these context cues, we compute the union region $\mathbf{b}_u \in \mathbb{R}^4$ for each human proposal $\mathbf{b}_h$ and object proposal $\mathbf{b}_o$,
and extract the corresponding appearance feature $v_u \in \mathbb{R}^D$ via RoI-align over the feature map $\bm{\Gamma}$ and self-attention aggregation. Similar union scores $s_u \in \mathbb{R}^N$ are computed with HOI embedding: $s_u = \textrm{Softmax}(\mathcal{W}_T \times v_u)$.

\subsubsection{Human-object Branch}\label{sec:pcn}
The human-object branch performs a fine-level classification for interaction pairs. For each human proposal $\mathbf{b}_h$ and object proposal $\mathbf{b}_o$, we crop the feature maps from $\bm{\Gamma}$ using RoI-Align, followed by a self-attention operation to generate their
appearance features $v_h, v_o \in \mathbb{R}^D$. We also compute a spatial feature $v_{sp}$ by encoding the relative positions of their bounding boxes $(\mathbf{b}_h, \mathbf{b}_o)$\footnote{For details c.f. Appendix}.
The holistic HOI representation $v_{ho} \in \mathbb{R}^{D} $ is an embedding of the human and object appearance features and their spatial feature: $v_{ho} = \mathcal{F}_{ho}([v_h; v_o; v_{sp}])$,
where $[;]$ is the concatenation operation and $\mathcal{F}_{ho}$ is a multi-layer perceptron (MLP). Finally, we use the shared HOI embedding to predict the pairwise interaction scores $s_{ho} \in \mathbb{R}^N$ for each human-object combination: $s_{ho} = \mathcal{W}_T \times v_{ho}$

\subsection{Model Learning with CLIP Supervision}\label{sec:learning}
In this section, we first utilize CLIP to generate two types of HOI supervision that are based on global images and local union regions, and then design a multi-task loss on various levels to train our multi-branch network. The overall loss function $\mathcal{L}$ consists of three terms: i) an image-wise recognition loss $\mathcal{L}_g$ to detect global HOIs; ii) a union loss $\mathcal{L}_u$ for identifying contextual regional HOIs;
and iii)
a pairwise interaction classification loss $\mathcal{L}_{ho}$ to guide the learning of instance-specific HOIs. Formally, the overall loss is written as: $    \mathcal{L} = \mathcal{L}_g + \mathcal{L}_u + \mathcal{L}_{ho}$.

\vspace{-1mm}
\paragraph{CLIP Supervision Generation}\label{sec:clip_supervision}
To obtain supervision for model learning, we directly employ CLIP on the image and union regions to generate the corresponding scores on the training set. As shown in Fig.~\ref{fig:overview}(b), 
we crop the image to a square region at its center, with the side length equal to its shortest edge.
This region is subsequently resized to $224\times 224$ pixels and fed into the pre-trained CLIP model to generate the global supervision $d_g \in \mathbb{R}^N$. Similarly, for each union box $\bm{b}_u$, we crop the corresponding region in the raw image and resize it to $224\times 224$ pixels, and then apply CLIP to generate local union supervision $d_u \in \mathbb{R}^N$.

\vspace{-1mm}
\paragraph{Image-wise loss $\mathcal{L}_g$:} Given the global image HOI scores $s_g$ and CLIP supervision $d_g$, $\mathcal{L}_g$ is 
a standard Kullback-Leibler (KL) divergence defined as: $\mathcal{L}_g = \mathcal{D}_{KL}(s_g || d_g)$. Notably, $d_g$ and $s_g$ are independent predictions from the original CLIP and the global branch, respectively. The aim of $\mathcal{L}_g$ is to align the up-scaled image feature map with the original CLIP representation, which plays a crucial role in extracting regional features for the union and human-object branches.

\vspace{-1mm}
\paragraph{Union loss $\mathcal{L}_u$:} Similarly, we take a KL divergence on union HOI scores $s_u$ and the local union supervision $d_u$ to formulate the union loss: $\mathcal{L}_u = \frac{1}{M} \sum_{m=1}^M \mathcal{D}_{KL}(s_u^m || d_u^m)$. Here $M$ is the total number of human-object combinations for a given image. 

\vspace{-1mm}
\paragraph{Human-object pairwise loss $\mathcal{L}_{ho}$:} 
Inspired by ~\cite{wang2020maf,wan2023weakly}, we adopt a Multiple Instance Learning (MIL) strategy to train the human-object branch.
In detail, we first string together all the interaction scores to form a bag, denoted as $S_{ho} = [s_{ho}^1;...;s_{ho}^M] \in \mathbb{R}^{M\cdot N}$, where $s_{ho}^m$ stands for the score of the $m$-th pair. Then we take maximization over all pairs to obtain the image-wise interaction scores: $\hat{s}_{ho} = \mathop{max}\limits_{m} S_{ho}$. This step essentially distills the most representative HOIs from the pairwise predictions, providing a consolidated representation of image-wise interaction scores that can be guided by the global CLIP supervision $d_g$.
Formally, the human-object loss $\mathcal{L}_{ho}$ is a KL divergence defined on $\hat{s}_{ho}$ and $d_g$: $\mathcal{L}_{ho} = \mathcal{D}_{KL}(\hat{s}_{ho}, d_g)$.

\subsection{Inference}\label{sec:inference}
During the inference stage, we combine multiple scores to obtain the final interaction score $s_{h,o}^r$ for each human-object pair $(\mathbf{b}_h,\mathbf{b}_o)$.  It includes the global HOI scores $s_g$, the union score $s_u$, the normalized pairwise interaction scores $p_{ho}$ (rather than $s_{ho}$), and the object detection scores $\langle s_h, s_o \rangle$ as follows:
\begin{equation}
    s_{h,o}^r = s_g \cdot s_u \cdot p_{ho} \cdot (s_h \cdot s_o)^{\gamma} 
    \label{eq:inference}
\end{equation}

where $\gamma$ is a hyper-parameter to balance the HOI scores and the object detection scores. 

It is noteworthy that we avoid using the original pairwise interaction score $s_{ho}$ as it fails to measure the contribution of each pair when multiple pairs in an image share the same interaction. Instead, we institute a competitive environment among the pairs by applying a $\textrm{Softmax}$ operation on $S_{ho}$:
$\bar{S}_{ho} = \mathop{\textrm{Softmax}} \limits_{m}(S_{ho})$. Following this, we derive the normalized pairwise interaction scores $p_{ho} = \sigma(\hat{s}_{ho}) \cdot \bar{s}_{ho}$, where $\bar{s}_{ho}$ is a row from $\bar{S}_{ho}$ and $\sigma$ is $\textrm{Sigmoid}$ function.
\section{Experiments}

\begin{table*}
\centering
\caption{\textbf{Results comparison of different methods on HICO-DET test set} (a full table of results comparison c.f. Appendix). \dag means re-implementation in~\cite{wan2023weakly}. Here FS, WS, and ZS indicate fully-supervised, weakly-supervised, and zero-shot HOI detection methods, respectively. The notation (D) means the visual encoder or the detector is pre-trained on dataset D, D$\in\{$COCO, HICO-DET, YFCC-15M$\}$.}
\resizebox{0.9\textwidth}{!}{
    \begin{tabular}{l|lll|ccc}
	\hline
	\multirow{2}{*}{S} & \multirow{2}{*}{Methods}  &  \multirow{2}{*}{Visual Encoder}&      \multirow{2}{*}{Detector}       & \multicolumn{3}{c}{HICO-DET (\%)} \\
			&&&& Full & Rare & Non-Rare \\
        \midrule
        \multirow{8}{*}{\rotatebox{90}{\textit{\textbf{FS}}}} &
	InteractNet~\cite{gkioxari2018detecting}  & RN50-FPN (COCO)  & FRCNN (COCO)        & 9.94 &7.16 & 10.77 \\
        & iCAN~\cite{gao2018ican}  & RN50 (COCO)     & FRCNN (COCO)        & 14.84 & 10.45 & 16.15  \\
        
        & TIN~\cite{li2019transferable}    & RN50-FPN (COCO) & FRCNN (COCO)  & 17.22 &13.51 & 18.32 \\
        
        & PMFNet~\cite{Wan_2019_ICCV}   & RN50-FPN (COCO)  & FRCNN (COCO)        & 17.46 &15.56 & 18.00 \\

	& HOTR ~\cite{kim_2021_CVPR}                   & RN50+Transformer (COCO) &DETR (HICO-DET)         &25.10 &17.34  & 27.42 \\
	& QPIC ~\cite{tamura2021qpic}                  & RN101+Transformer (COCO) &DETR (COCO)            &29.90 &23.92  & 31.69 \\          
	& GEN-VLKT ~\cite{liao2022gen}  & RN50+Transformer (HICO-DET)      &DETR (HICO-DET)  & 33.75 & 29.25 & 35.10 \\
 	& HOICLIP ~\cite{ning2023hoiclip}                 & RN50+Transformer (HICO-DET)      &DETR (HICO-DET)       &34.69   &31.12 &35.74 \\
        \midrule
        \multirow{4}{*}{\rotatebox{90}{\textit{\textbf{WS}}}} &
    Explanation-HOI\dag ~\cite{baldassarre2020explanation}  &ResNeXt101 (COCO)     &FRCNN (COCO)  & 10.63 &8.71 & 11.20 \\
    & MX-HOI    ~\cite{kumaraswamy2021detecting}           &RN101 (COCO)             &FRCNN (COCO)  & 16.14 &12.06&17.50 \\           
    & PPR-FCN\dag  ~\cite{zhang2017ppr}                        &RN50  (YFCC-15M)      &FRCNN (COCO)  & 17.55 &15.69&18.41 \\
    & PGBL~\cite{wan2023weakly}                        &RN50 (YFCC-15M)                &FRCNN (COCO)   & 22.89  & 22.41 & 23.03\\
        \midrule
        \multirow{2}{*}{\rotatebox{90}{\textit{\textbf{ZS}}}} &
        \textit{baseline}  & RN50 (YFCC-15M)                &FRCNN (COCO)   & 10.48  & 9.45 & 10.78 \\
        & \cellcolor{yellow!30} \textit{ours}  & \cellcolor{yellow!30} RN50 (YFCC-15M)  & \cellcolor{yellow!30} FRCNN (COCO)   & \cellcolor{yellow!30} 17.12  & \cellcolor{yellow!30} 20.26 & \cellcolor{yellow!30} 16.18 \\
        \hline
    \end{tabular}
    }
    \vspace{-3mm}
    \label{hoi-full}
\end{table*}

\subsection{Datasets and Metrics}\label{setting}
We use the public HOI detection dataset HICO-DET to benchmark our model. The dataset contains 37,633 training images and 9,546 test images. 
It includes $C=80$ common objects (the same as MSCOCO~\cite{lin2014microsoft}) and 117 unique action categories, together forming $N=600$ HOI categories.

We adopt the mean average precision (mAP) metric~\cite{chao:iccv2015} for evaluating HOI detection results. A human-object pair is deemed positive when the predicted human and object boxes have an IoU of at least 0.5 with their ground truth boxes, and the HOI class is correctly classified.

\subsection{Implementation Details}\label{sec:imp_detail}

We use an off-the-shelf Faster R-CNN~\cite{ren2015faster} pre-trained on MSCOCO to generate up to 100 object candidates for each image. It's crucial to note that we only keep the detection results and do not re-use the feature maps. We rather employ a CLIP with a ResNet-50 visual encoder, which is pre-trained on the YFCC-15M dataset~\cite{yfcc}, with an image resolution of $224\times 224$. The CLIP model we used was implemented by OpenCLIP~\footnote{\url{https://github.com/mlfoundations/open_clip}}, which achieved an mAP of 35.5 on zero-shot image-wise HOI recognition on the test set, suggesting that it is capable of providing a comprehensive holistic understanding of HOIs.

During training, we use a larger image resolution with a minimum edge length of 384, while maintaining the original aspect ratio of the input images in our multi-branch network. 
We freeze the weights of the pre-trained visual encoder and optimize the remaining modules by AdamW, with a learning rate of 1e-4 and batch size of 16.
The model is trained for 30K iterations on two NVIDIA V100 GPUs. After 15K iterations, the learning rate is decayed by a factor of 10.  Following previous works~\cite{zhang2021spatially,li2019transferable}, we set feature dimension $D$ as 1024 and the detection score weight $\gamma$ as 2.8.

\subsection{Quantitative Results} \label{quant-results}
As shown in Tab.~\ref{hoi-full}, with our multi-level knowledge integration strategy from CLIP, our approach achieves 17.12 mAP with ResNet-50, which is on par or even surpasses some fully-supervised (FS)~\cite{gkioxari2018detecting,gao2018ican,li2019transferable,Wan_2019_ICCV} and weaky-supervised (WS) ~\cite{baldassarre2020explanation,kumaraswamy2021detecting, zhang2017ppr} methods. For the FS and WS methods, we observe the mAP on Non-rare classes (i.e., those with more than 10 HOI instance annotations in the training set) is always higher than on Rare classes. 
This skew is to be expected given that HICO-DET is inherently an imbalanced dataset~\cite{shen2018scaling,gupta2019nofrills}. Models tend to learn frequently occurring patterns for which they have training supervision. Despite certain HOIs being simpler to learn, their performance may lag due to the relative lack of supervision, compared to the more challenging yet annotated HOIs.

Remarkably, we obtain a higher mAP on Rare classes compared to Non-Rare classes in our results. 
This consequence stems from the integration of CLIP for HOI representation learning and model supervision. 
Firstly, CLIP is pre-trained on large-scale image-text pairs and has potentially encountered every imaginable HOI scenario during its pre-training phase.
We build our model on top of CLIP components, which allows us to exploit its strong generalization capability for learning a better HOI representation.
Secondly, when comparing our results with PGBL~\cite{wan2023weakly}, which also exploits CLIP for HOI representation learning and achieves the best performance in a weakly-supervised setting (i.e., image-level HOI annotations are available). We experience a small drop on Rare classes (from $22.41 \xrightarrow{} 20.26$), but a significant drop on Non-Rare classes (from $23.03 \xrightarrow{} 16.18$).
This disparity suggests that the learning of Non-Rare HOIs is more reliant on strong annotations, whereas Rare HOIs can be effectively learned by distilling the `dark knowledge' from CLIP scores. Consequently, we sidestep issues associated with the long-tailed distribution.

\subsection{Ablation Studies}\label{ablation}
In this section, we mainly assess the effectiveness of each component with detailed ablation studies on HICO-DET dataset. We first introduce our baseline, based on which we will answer some interesting questions regarding the model design and learning.

\vspace{-4mm}
\paragraph{Baseline:} 
Our baseline model is constructed on top of the human-object branch, where the human-object representation $v_{ho}$ is used to predict their normalized interaction scores $s_{ho}$, which are supervised by $\mathcal{L}g$. During the inference process, the final interaction scores in ~\ref{eq:inference} are recomputed as $s_{h,o}^r = p_{ho} \cdot (s_h \cdot s_o)^{\gamma}$.

\begin{table}[t]
	\centering
	\caption{\textbf{Comparison of inference speed and performance} between baseline, training-free (TF) approach, and our method. 
    TF* means $d_g$ is added to predict $s^r_{ho}$ on top of TF.
 }
 \vspace{-2mm}
	\resizebox{0.38\textwidth}{!}{
		\begin{tabular}{c|c|ccc}
			\hline
			\multirow{2}{*}{Exp}  & \multirow{2}{*}{Speed (fps)}  & \multicolumn{3}{|c}{mAP (\%)}\\
			             &  & Full & Rare & Non-Rare\\
	\hline
        \textit{base} & \textbf{56.19} & 10.48 & 9.45 & 10.78  \\
        \textit{TF}  & 6.52 & 11.19 & 13.98 & 10.37   \\
        \textit{TF}*  & 6.47 &12.24 & 15.75 & 11.19   \\
        \textit{ours}   & 35.64 & \textbf{17.12} & \textbf{20.26} & \textbf{16.18}  \\
        \hline 
	\end{tabular}}
        \vspace{-2mm}
	\label{hico-aba1}
\end{table}

\begin{table}[t]
	\centering
	\caption{\textbf{Ablation study of multi-level incorporation on HICO-DET dataset.} The baseline is the human-object (h-o) branch, and we add other branches on top of it. We denote 'early' as the early fusion of union features and 'late' as the late fusion of union scores.}
    \vspace{-2mm}
	\resizebox{0.49\textwidth}{!}{
		\begin{tabular}{c|ccc|ccc}
			\hline
			\multirow{2}{*}{Exp}  & \multicolumn{3}{c|}{Branch}  & \multicolumn{3}{|c}{mAP (\%)}\\
			             & h-o & union & global & Full & Rare & Non-Rare\\
	\hline
        \textit{0} & \checkmark & - & - & 10.48 & 9.45 & 10.78  \\
        \textit{1}   & \checkmark & \checkmark(late) & - &  14.49  & 16.33 & 13.95  \\
        \textit{2}   & \checkmark & - & \checkmark & 15.84  & 17.91 & 15.21  \\
        \textit{3}   &\checkmark & \checkmark(early) & \checkmark & 14.64 & 16.09 & 14.21 \\
        \textit{4}   &\checkmark & \checkmark(late) & \checkmark & \textbf{17.12} & \textbf{20.26} & \textbf{16.18} \\
        \hline
	\end{tabular}}
        \vspace{-2mm}
	\label{hico-aba2}
\end{table}

\vspace{-4mm}
\paragraph{Why not a training-free (TF) approach?} 
In a TF approach, we directly apply CLIP scores on union regions $d_u$ along with object detection scores $\langle s_h, s_o \rangle$ for inference. This reformulates Eq.~\ref{eq:inference} as: $s_{h,o}^r=d_u\cdot (s_h \cdot s_o)^{\gamma}$.


While this straightforward approach only relies on a pre-trained CLIP and avoids the need for model training, it has two notable drawbacks compared to our method:
(i) It is time-consuming for inference. For each image, the approach requires forwarding the CLIP model for the image and all union regions, resulting in $M+1$ forward passes. As indicated in Tab.~\ref{hico-aba1}, given the detected bounding boxes, the inference speed for the TF approach is only 6.52 frames per second (fps) on a single Nvidia V100 GPU with a batch size of 1, whereas our method achieves an fps of 35.64.
(ii) The performance is low, even enhanced by $d_g$ (i.e., TF*). This issue arises because both $d_g$ and $d_u$ correspond to large regions, and as such, they fail to specify the interacted human and object pair. Consequently, the CLIP scores, when directly used for inference, tend to be noisy. In contrast, our method opts to distill knowledge from these scores at different levels, leading to a significant performance increase, from an mAP of 12.24 to 17.12.

\vspace{-2mm}
\paragraph{Does multi-level CLIP knowledge distillation strategy work?} 
As demonstrated in Tab.~\ref{hico-aba2}, the answer is definitively yes. We added the union branch and the global branch on top of our baseline model (Exp 0). The results indicate that the union branch enhances the mAP from 10.48 to 14.49 (Exp 0 vs. 1), while the global branch raises the mAP from 10.48 to 15.84 (Exp 0 vs. 2). When we combine both branches, the mAP sees a more significant improvement, from 10.48 to 17.12 (Exp 0 vs. 4). These results underscore the effectiveness of the multi-level CLIP knowledge distillation strategy.

\begin{table}[t]
	\centering
	\caption{\textbf{Ablation study of different supervision strategies on HICO-DET dataset.} \textbf{g} means supervision from $d_g$ and \textbf{u} means supervision from union region CLIP scores $d_u$. \textbf{g+u} indicates using both supervisions. In this table, the global branch is added for all the experiments and supervised with $d_g$. 
 }
 \vspace{-2mm}
	\resizebox{0.4\textwidth}{!}{
		\begin{tabular}{c|cc|ccc}
			\hline
			\multirow{2}{*}{Exp}  & \multicolumn{2}{c|}{Branch}  & \multicolumn{3}{|c}{mAP (\%)}\\
			             & union & h-o & Full & Rare & Non-Rare\\
	\hline
        \textit{0} &g &g & 15.05 & 16.76 & 14.55  \\
        \textit{1}   &u &u & 14.22 & 15.00 & 13.99  \\
        \textit{2}   &u &g & \textbf{17.12} & \textbf{20.26} & \textbf{16.18}  \\
        \textit{3}   &u &g+u & 15.83 & 18.20 & 15.13  \\
        \textit{4}   &g+u &g & 16.96 & 19.54 & 16.18  \\
        \hline 
	\end{tabular}}
	\label{hico-aba3}
        \vspace{-2mm}
\end{table}

\begin{table}[t]
	\centering
	\caption{\textbf{Generalization to Unseen Categories}. We randomly select $N'$ HOI categories for model learning.
 }
 \vspace{-2mm}
	\resizebox{0.27\textwidth}{!}{
		\begin{tabular}{c|ccc}
			\hline
			\multirow{2}{*}{$N'$}  & \multicolumn{3}{|c}{mAP (\%)}\\
			             & Full & Rare & Non-Rare\\
	\hline
        100 & 15.25 & 18.97 & 14.15  \\
        300 & 16.45 & 19.48 & 15.55  \\
        600 & \textbf{17.12} & \textbf{20.26} & \textbf{16.18}  \\
        \hline 
	\end{tabular}}
        \vspace{-3mm}
	\label{hico-aba4}
\end{table}

\begin{figure*}
    \centering
    \includegraphics[width=0.9\linewidth]{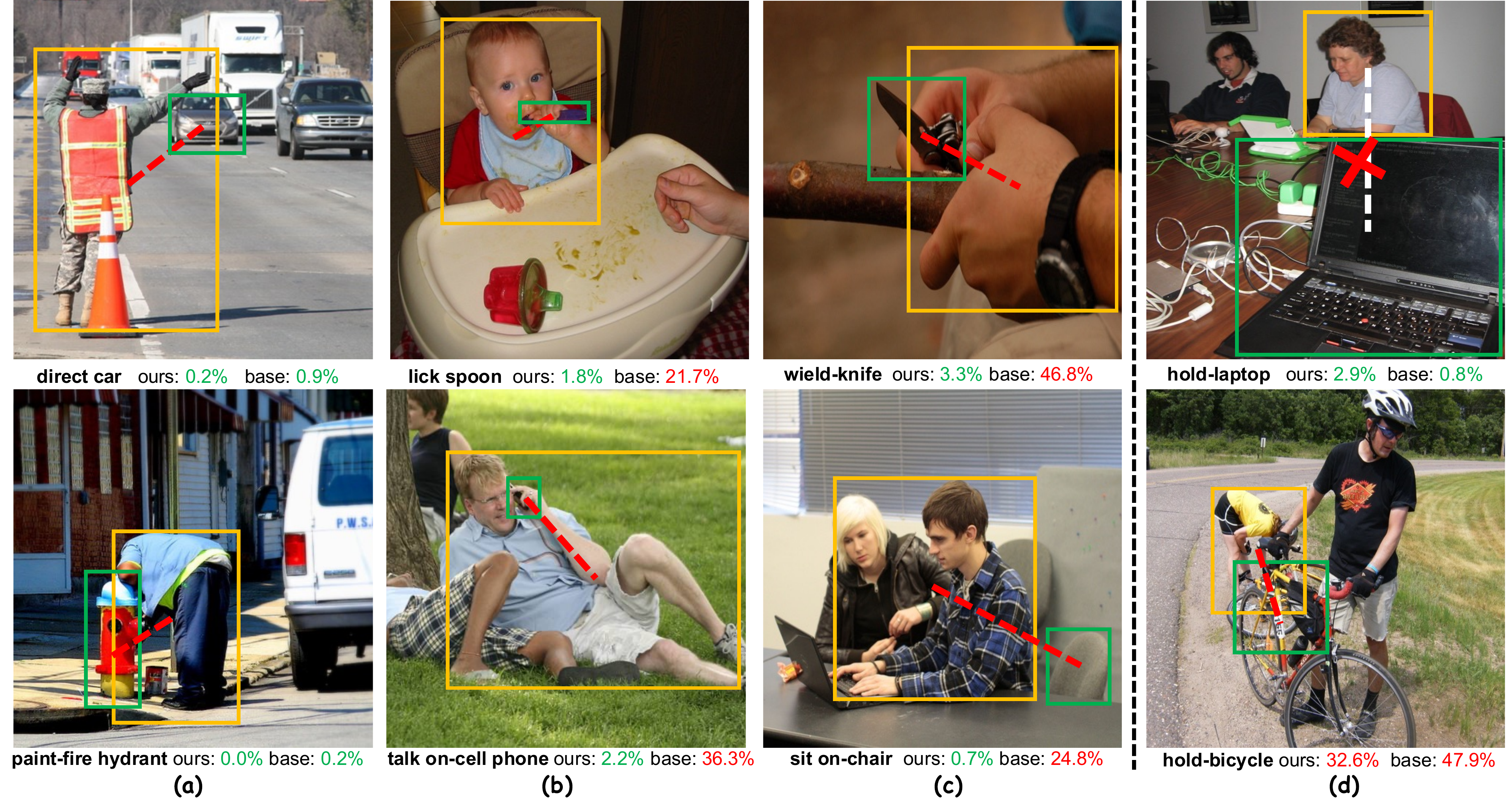}
    \vspace{-4mm}
    \caption{\textbf{Visualization of the HOI detection results.}
    We compare our method with the baseline model based on the relative ranking of the detection scores, which is presented in percentile format.
    The percentiles highlighted in \textcolor[RGB]{0,192,0}{green} signify the model's confident HOI predictions, whereas those in \textcolor[RGB]{255,0,0}{red} indicate negative HOI predictions that the model treats as background.
    }
    \vspace{-4mm}
    \label{fig:visulization}
\end{figure*}

\vspace{-2mm}
\paragraph{How to incorporate contextual cues from the union region?} In Tab.~\ref{hico-aba2}, we also explore two different designs for integrating the contextual union branch, including an early fusion strategy and a late fusion strategy. For the early fusion strategy, we concatenate the union feature $v_u$ with human-object appearance features $\langle v_h,v_o \rangle$ and their spatial encoding $v_{sp}$ to compose $v_{ho}$. The experimental results (Exp 3 vs. 4) indicate that the early fusion strategy performs considerably worse than a late fusion strategy (14.64 vs 17.12), and even underperforms Exp 2 where the union branch is not included. 

We hypothesize that this is because the union feature, encompassing a large region, may include other HOIs or background distractions, making it noisy. These irrelevant features may not offer valuable contextual information for HOI representation learning, particularly in a zero-shot setup where the training signals are also somewhat noisy. Conversely, with the late fusion strategy, the union branch aims to learn context-aware union HOI scores. Although these union scores cannot precisely describe the specific human-object pair, they provide some contextual HOI cues that are compatible with the HOI predictions from the other branches.

\vspace{-2mm}
\paragraph{What kind of supervision works best for different branches?} In Tab.~\ref{hico-aba3}, we compare different supervision strategies for each branch. By default, we supervise the image branch with global supervision $d_g$. Then, we apply $d_g$ on the union branch in the same manner as $\mathcal{L}_{ho}$ in Exp 0. This results in an mAP drop from 17.12 to 15.05, indicating that the union features, being noisy and non-discriminative, are not suitable for a MIL strategy.
Besides, we apply local supervision $d_u$ on the human-object branch in Exp 1 and observe the drop in mAP to 14.22. This is because of the noisiness of $d_u$, as it does not always correspond accurately to a specific human-object pair.

Experimental results reveal that the mismatch of the supervision signals can result in a performance drop, and  
Combining both types of supervision does not necessarily lead to better results (as seen in Exp 3 and 4).

\vspace{-3mm}
\paragraph{Can our method generalize to unseen categories?}
To answer this question, we randomly select $N'$ out of $N=600$ HOI categories on HICO-DET during training. This implies the class dimensions of predicted scores $\langle s_g, s_u, s_{ho} \rangle$ and CLIP supervisions $d_g, d_u$ are set to $N'$. For inference, we evaluate across all 600 categories.
As shown in Tab.~\ref{hico-aba4}, even though the training is limited to just 100 categories, we observed only a slight $1.87$ mAP drop compared to the final model. This indicates that our approach effectively captures 
common knowledge that can be shared across all HOI categories.

\subsection{Qualitative Results}
Figure \ref{fig:visulization} offers a qualitative evaluation of our method. Due to the way we incorporate multiple scores into our final prediction $s_{ho}^r$ in our methodology, it's not practical to directly compare the HOI scores with the baseline.
Instead, our visualization and comparison with the baseline model are based on the relative ranking of the detection scores rather than their absolute values. Concretely, for each HOI prediction, we present its ranking amongst all predictions belonging to the same category across the entire test set. 
The ranking is exhibited in the form of a percentile (top $p\%$), wherein a lower percentile value (smaller $p$) signifies the model's strong confidence in the positive prediction (represented by \textcolor[RGB]{0,192,0}{green} colored numbers).

As depicted in Fig.\ref{fig:visulization}(a), both our method and the baseline successfully identify some Rare HOI classes. However, when the objects are quite small, as shown in Fig.\ref{fig:visulization}(b), our model tends to offer more confident predictions, attributing this advantage to its ability to factor in contextual cues. For example, in the top image, our model infers that a baby is likely licking a spoon due to the context of sitting in a baby chair with residual sauce visible.
In a similar vein, Fig.\ref{fig:visulization}(c) showcases situations where objects are heavily occluded. Despite this challenge, our model manages to discern some unapparent relationships by considering the overall environment. As an illustration, in the image at the bottom, a man in an office, engaged in computer work on the table, is more likely identified by our model as 'sitting on a chair', which would be a challenging task for the baseline method.

\vspace{-2mm}
\section{Limitations and Future Works}
\vspace{-1mm}
Although inspiring results have been achieved by our method, the zero-shot HOI detection is far from satisfactory. 
As an example of its limitations, the top image in Fig.\ref{fig:visulization}(d) shows a typical failure case where our model incorrectly associates a person with a computer that is considerably distant. This error can be attributed to the absence of adequate supervision for pairwise associations. 
Furthermore, the bottom image exhibits a scenario where our method struggles to recognize relations when the object is completely obscured.

A potential area for further exploration based on this work involves the detection of ambiguous HOI associations. Previous research has investigated this issue within fully-supervised or weakly-supervised settings~\cite{li2018transferable,liu2022interactiveness,wan2023weakly}. However, transferring these learnings to a zero-shot setup remains a largely unexplored area. 
Besides, this study employs a classic CLIP structure for zero-shot HOI detection. Nonetheless, it is intriguing to explore various adaptations of CLIP~\cite{zhai2022lit,yao2022filip,zhai2023sigmoid} for enhancing performance in this task.

\vspace{-2mm}
\section*{Acknowledgement}
\vspace{-2mm}
We acknowledge funding from European Research Council under the European Union’s Horizon 2020 research and innovation program (Grant Agreement No.
101021347) and Flemish Government under the Onderzoeksprogramma Artificiele
Intelligentie (AI) Vlaanderen programme.

{\small
\bibliographystyle{ieee_fullname}
\bibliography{egbib}
}

\end{document}